\documentclass[11pt]{article}
\usepackage{style/emnlp2016}
\usepackage{times}
\usepackage{enumitem}
\setlist{nosep,after=\vspace{6.0pt}}

\emnlpfinalcopy
\newif\ifcomment\commentfalse
\usepackage{mdwlist}
\usepackage{latexsym}
\usepackage{nicefrac}
\usepackage{booktabs}
\usepackage{amsfonts}
\usepackage{bold-extra}
\usepackage{amsmath}
\usepackage{amssymb}
\usepackage{bm}
\usepackage{xcolor}
\usepackage{comment}
\usepackage{tabularx}
\usepackage{environ}
\usepackage{dsfont}
\usepackage{textcomp}
\usepackage{float}
\usepackage{graphicx}
\usepackage{mathtools}
\usepackage{microtype}
\usepackage{multirow}
\usepackage{multicol}
\usepackage{url}
\usepackage{latexsym}
\usepackage{todonotes}

\newcommand{\gem}[1]{\mbox{\textsc{gem}}}
\newcommand{\abr}[1]{\textsc{#1}}

\newcommand{\hidetext}[1]{}
\newcommand{\ignore}[1]{}

\ifcomment

\fi

\newcommand{\smallurl}[1]{ \begin{scriptsize}\url{#1}\end{scriptsize}}

\NewEnviron{smalign}{
\vspace{-.6cm}
\begin{small}
\begin{align}
  \BODY
\end{align}
\end{small}
\vspace{-.6cm}
}

\definecolor{CUgold}{HTML}{CFB87C}
\definecolor{grey}{rgb}{0.95,0.95,0.95}
\definecolor{ceil}{rgb}{0.57, 0.63, 0.81} 
\newcolumntype{L}{>{\arraybackslash}m{6.5cm}}

\newcommand{\wtq}[0]{WTQ}
\newcommand{\name}[0]{SQA}
\newcommand{\fp}[0]{{\bf \textsc{fp}}}
\newcommand{\neural}[0]{{\bf \textsc{neural}}}
\newcommand{\bmat}[1]{\text{\textbf{#1}}}
\newcommand{\bvec}[1]{\boldsymbol{#1}}

\title{Answering Complicated Question Intents \\Expressed in Decomposed Question Sequences}

\author{
Mohit Iyyer\thanks{\hspace{.06in}Work done during an internship at Microsoft Research}\\
Department of Computer Science and \abr{umiacs}\\
University of Maryland, College Park\\
{\tt miyyer@umd.edu} \And
Wen-tau Yih, Ming-Wei Chang\\
Microsoft Research\\
Redmond, WA 98052  \\
{\tt \{scottyih,minchang\}@microsoft.com}
}

\begin{document}
\maketitle

\begin{abstract}
Recent work in semantic parsing for question answering has focused on long and complicated questions, many of which would seem unnatural if asked in a normal conversation between two humans. In an effort to explore a conversational QA setting, we present a more realistic task: answering sequences of simple but inter-related questions. 
We collect a dataset of 6,066 question sequences that inquire about semi-structured tables from Wikipedia, with 17,553 question-answer pairs in total. 
%We collect a dataset of question sequences by asking crowdsourced workers to decompose questions from WikiTableQuestions, which contains highly-compositional questions about tables from Wikipedia.
Existing QA systems face two major problems when evaluated on our dataset: (1) handling questions that contain coreferences to previous questions or answers, and (2) matching words or phrases in a question to corresponding entries in the associated table. We conclude by proposing strategies to handle both of these issues.
\end{abstract}

\section{Introduction}
\label{sec:intro}

Semantic parsing, which maps natural language text to meaning representations in formal logic, has emerged as a
key technical component for building question answering systems~\cite{liang2016}. Once a natural language question has been mapped to a formal query, its answer can be retrieved simply by executing the query on a back-end structured database.

% As a result, the correctness of the answers is typically used for measuring the quality of the semantic parsers.\todo{mohit:this last sentence feels not related to the scope of the paper}

One of the main focuses of semantic parsing research is how to address \emph{compositionality} in language.
Extremely complicated questions have been used to demonstrate the sophistication of semantic parsers,\footnote{For example,
``\emph{will it be warmer than 70 degrees near the Golden Gate Bridge after 5PM the day after tomorrow?}"~\cite{viv}} and such questions have been specifically
targeted in the design of a recently-released QA dataset~\cite{pasupat2015compositional}. Take for example the following question: %, given a movie database:
``\emph{of those actresses who won a Tony after 1960, which one took the most amount of years after winning the Tony to win an Oscar?}''
The corresponding logical form is highly compositional; in order to answer it, many sub-questions must be implicitly answered in the process (e.g., \emph{``who won a Tony after 1960?''}).

While we agree that semantic parsers \emph{should} be able to answer very complicated questions, in reality these questions are rarely issued by human users of QA systems.\footnote{As indirect evidence, the percentage of questions with more than 15 words is only 3.75\%
in the WikiAnswers questions dataset~\cite{fader2014open}.}
Because users can interact with a QA system repeatedly,
there is no need to assume a single-turn QA setting where the exact \emph{question intent} has to be captured with just one complex question.  % SY: making it sound more formal
%why should we assume a single-turn QA setting where the exact \emph{question intent} has to be captured with just one complex question?
The same intent can be more naturally expressed through a sequence of simpler
questions, as shown below:

\begin{enumerate}
	\item \emph{What actresses won a Tony after 1960?}
	\item \emph{Of those, who later won an Oscar?}
	\item \emph{Who had the biggest gap between their two award wins?}
\end{enumerate}
Decomposing complicated intents into multiple related but simpler questions is arguably
a more effective strategy to explore a topic of interest, and it reduces the cognitive burden on both the person who asks the question as well as the one who answers it.\footnote{While cognitive load has not been measured specifically for complicated questions, there have been many studies linking increased sentence complexity to longer reading times~\cite{hale2006uncertainty,levy2008expectation,frank2013uncertainty}.}

%---------------------------------------------------------------------------------------------------------------------------------------------------------

In this work, we study the semantic parsing problem for answering \emph{sequences} of simple related questions.
We collect a dataset of question sequences that we call SequentialQA (\name)\footnote{To be released at \url{http://aka.ms/sqa}} by asking crowdsourced workers to decompose complicated
questions sampled from the WikiTableQuestions dataset~\cite{pasupat2015compositional} into multiple easier ones. In addition, each question is
associated with answers selected by workers from a corresponding Wikipedia HTML table.
Using the \name~dataset, we investigate experimentally how we should modify traditional semantic parser design to address different properties in this
new, multi-turn QA setting, such as inter-question coreferences.
%---------------------------------------------------------------------------------------------------------------------------------------------------------

% Main contributions

%\todo{SY: need to be careful whenever claiming the first; checked Percy Liang's CACM survey paper}
Our contributions are twofold. First, to the best of our knowledge, \name\ is the first semantic parsing dataset for sequential question answering.
We believe this dataset will be valuable to future research on both semantic parsing and question answering in the more natural interactive setting.
Second, after evaluating existing question answering systems on \name, we find that none of them performs adequately, despite the relative lack of compositionality.
We provide a detailed error analysis,
which suggests that improperly-resolved references and mismatches between question text and table entries are the main sources of errors.

The rest of the paper is structured as follows. Sec.~\ref{sec:related} contrasts the existing tasks and datasets to \name. Sec.~\ref{sec:dataset}
describes how we collect the data in detail. Sec.~\ref{sec:experiments} presents our experimental study, followed by the discussion in Sec.~\ref{sec:discussion}.
Finally, Sec.~\ref{sec:conclusion} concludes the paper.
%\todo{mohit: is this last paragraph necessary? \\ SY: A formality thing, I know some people really insist that every intro should ends with a paragraph like this.  We can remove it for a short paper submission if we need space.}

\section{Related Work}
\label{sec:related}

Our work is related to existing research on conversational (or contextual) semantic parsing, as well as more generally to interactive question-answering systems that operate on semi-structured data.

Previous work on conversational QA has focused on small, single-domain datasets. Perhaps most related to our task is the context-dependent sentence analysis described in ~\newcite{zettlemoyer2009learning}, where conversations between customers and travel agents are mapped to logical forms after resolving referential expressions. Another dataset of travel booking conversations is used by~\newcite{artzi2011bootstrapping} to learn a semantic parser for complicated queries given user clarifications. More recently,~\newcite{longpasupatliang2016} collect three contextual semantic parsing datasets (from synthetic domains) that contain coreferences to entities and actions. We differentiate ourselves from these prior works in two significant ways: first, our dataset is not restricted to a particular domain, which results in major challenges further detailed in Section~\ref{sec:discussion}; and second, a major goal of our work is to analyze the different types of sequence progressions people create when they are trying to express a complicated intent.

Complex, interactive QA tasks have also been proposed in the information retrieval community, where the data source is a corpus of newswire text~\cite{kelly2007overview}.
We also build on aspects of some existing interactive question-answering systems. For example, the system of~\newcite{harabagiu2005experiments} includes a module that predicts what a user will ask next given their current question. A follow-up work~\cite{lacatusu2006impact} proposes syntax-based heuristics to automatically decompose complex questions into simpler ones. Both works rely on proprietary limited-domain datasets; it is unlikely that the proposed heuristics would scale across arbitrary domains. 
\section{A Dataset of Question Sequences}
\label{sec:dataset}

\ignore{
An additional advantage of \wtq\ is that the entire knowledge base for a given question (the associated HTML table) can be easily displayed in a UI to workers, unlike datasets that rely on large knowledge bases such as Freebase. In the rest of this section, we detail the data collection process and offer statistics describing \name.
}
%
%We initially explored using WebQuestions~\cite{berant2014semantic} as a source for these intents and answers; however, WebQuestions contains primarily simple ``one-hop'' questions that cannot be further decomposed.

\begin{figure}[t]
\includegraphics[width=1.0\linewidth]{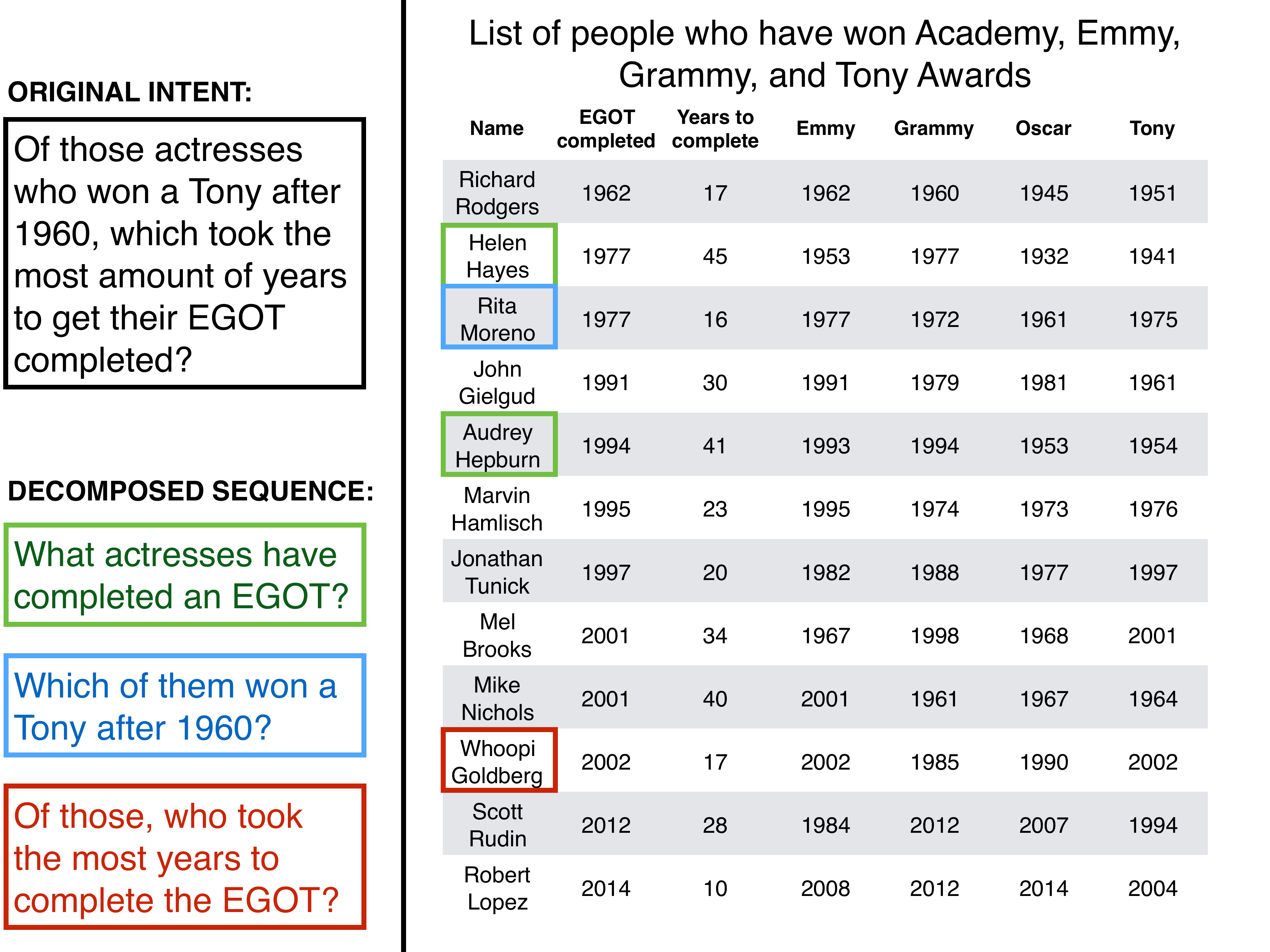}
  \caption{An example decomposition of a complicated intent from \wtq. Workers must create a sequence of decomposed questions where the answer to each question is a subset of cells in the table.}
\label{fig:decomposition}
\end{figure}

%\subsection{Selecting Intents for Decomposition}
%\label{ssec:decomposition_process}

Since there are no previous publicly-available datasets for our task, we collect the SequentialQA (\name) dataset via crowdsourcing.
We leverage WikiTableQuestions~\cite[henceforth \wtq]{pasupat2015compositional}, which contains highly compositional questions associated with HTML tables from Wikipedia.
Each crowdsourcing task contains a long, complex question originally from \wtq\ as the question \emph{intent}. The workers are asked to compose a sequence of simpler questions
that lead to the final intent; an example of this process is shown in Figure~\ref{fig:decomposition}.

To simplify the task for workers, we only select certain types of questions from \wtq. In particular, we only use questions from \wtq\ whose answers are cells in the table, which excludes those involving arithmetic and counting. We likewise also restrict the questions our workers can write to those that are answerable by only table cells. These restrictions speed the annotation process because, instead of typing their answers, workers can just click on the table to answer their question. They also allow us to collect answer coordinates (row and column in the table) as opposed to answer text, which removes many normalization issues for answer string matching that are present in the original \wtq\ dataset. Finally, we only use intents that contain nine or more words; we find that shorter questions tend to be simpler and are thus less amenable to decomposition.

%\subsection{Constraining the UI}

After iterating on the task design with many pilot tasks, we found that the following constraints are necessary for workers to produce good sequences:

\paragraph{Minimum sequence length:} Workers must create sequences that contain at least two questions. If the intent is not easily decomposed into multiple questions, we instruct workers to create an alternate intent whose answer is the same as that of the original. We also encourage workers to write longer sequences if possible.

\paragraph{Final answer same as original answer:} The final question of a sequence must have the same answer as that of the original intent. Without this constraint, some workers were writing sequences that diverged drastically from the intent.

\paragraph{No copying the intent:} After adding the previous constraint, we found that many workers were just copying the intent as the final question of their sequence, which resulted in unnatural-sounding sequences. After we disallowed copying, the workers' final questions contained many more references to previous questions and answers.
\\\\
We also encouraged (but did not enforce) the following:

\paragraph{Simplicity:} When decomposing a complicated intent into a sequence of questions, we expect that each question in the sequence should be simpler than the intent itself. However, defining ``simple'' is difficult, and enforcing any definition is even harder. Instead, we told workers to try to limit their questions to those that can be answered with just a single primitive operation (e.g., column selection, argmax/argmin, filtering with a single condition) and provided them with examples of each primitive. Following this definition too closely, however, can result in unnatural sequences, so we do not make any UI changes to limit questions to single primitives.

\paragraph{Inter-question coreferences:} Take the following two sequences generated from the same question intent:
\begin{enumerate}
	\item What country won the World Cup in 2014? Of the players on the team that won the World Cup in 2014, which ones were midfielders?
	\item What country won the World Cup in 2014? Of the players on that team, which ones were midfielders?
\end{enumerate}
The second question of the first sequence clumsily repeats information found in the preceding question, while the second sequence avoids this repetition with the referring expression ``that team''.
%
% \todo{SY: please confirm mohit:yes, we did}
To encourage more coreferences between questions, we showed workers example sequences like these and stated that the second one is preferred. 

%\subsection{\name\ Statistics}
\subsection{Properties of \name}
In total, we used 2,022 question intents from the train and test folds of the \wtq\ for decomposition.
%, selecting only those that meet the conditions described in Section~\ref{ssec:decomposition_process}.
We had three workers decompose each intent, resulting in 6,066 unique questions sequences containing 17,553 total question-answer pairs (for an average of 2.9 questions per sequence). We divide the dataset into train and test using the original \wtq\ folds, resulting in an 83/17 train/test split. Importantly, just like in \wtq, none of the tables in the test set are seen in the training set.

We identify three frequently-occurring question classes: \emph{select column}, \emph{select subset}, and \emph{select row}. In \emph{select column} questions, the answer is an entire column of the table; these questions account for 23\% of all questions in \name. Subset and row selection are more complicated than column selection, particularly because they usually contain coreferences to the previous question's answer. In \emph{select subset} questions, the answer is a subset of the previous question's answer; similarly, the answers to \emph{select row} questions occur in the same row(s) as the previous answer but in a different column. \emph{Select subset} questions make up 27\% of \name, while \emph{select row} is 19\%. The remaining 31\% of \name\ is comprised of more complex questions that are combinations of these three types. In the sequence ``\emph{what are all of the tournaments? in which one did he score the least points? on what date was that?}'', the first question is a column selection, the second question is a subset selection, and the final question is a row selection.

We also observe dramatic differences in the types of questions that are asked at each position of the sequence. For example, looking at just the first question of each sequence, 51\% of them are of the \emph{select column} variety (e.g., ``what are all of the teams?''). This number dwindles to just 18\% when we look at the second question of each sequence, which indicates that the collected sequences start with general questions and progress to more specific ones. By definition, \emph{select subset} and \emph{select row} questions cannot be the first question in a sequence.

\section{Baseline Experiments}
\label{sec:experiments}

We evaluate two existing QA systems on \name, a semantic parsing system called \emph{floating parser} and an \emph{end-to-end neural network}.  The floating parser considers each question in a sequence independently of the previous questions, while the neural network leverages contextual information from the sequence.
Our goals with these experiments are (1) to measure the difficulty of \name and (2) to better understand the behaviors of existing state-of-the-art systems.

\subsection{Floating parser}
An obvious baseline is the floating parser (\fp) developed by~\newcite{pasupat2015compositional}, which
\fp\ maps questions to logical forms and then executes them on the table to retrieve the answers.
It achieves 37.0\% accuracy on the \wtq\ test set.
One of the key challenges in semantic parsing is the ``semantic matching problem'', where question text cannot be matched to the corresponding answer column or cell verbatim.
Without external knowledge, it is often hard to map words or phrases in a question to predicates in its corresponding logical form. Further compounding this problem is that the train and test tables are disjoint, which renders lexicon induction futile. Therefore, \fp\ does not anchor predicates to tokens in the question, relying instead on typing constraints to reduce the search space.\footnote{See~\newcite{pasupat2015compositional} for more details.}

Using \fp\ as-is results in poor performance on \name. The main reason is that the system is configured for questions with single answers, while \name\ contains a high percentage of questions with multiple-cell answers. We address this issue by removing a pruning hyperparameter (\emph{tooManyValues} %\todo{SY: is this a feature? mohit: no, it's a hyperparameter of the system})
that eliminates all candidate parses with more than ten items in their denotations, as well as by removing features that add bias on the denotation size.

%We retrain the modified system using a beam size of 50; we find that increasing the beam size to the optimal \wtq\ setting of 200 does not improve accuracy on \name.

\subsection{End-to-end neural network}
Recently, two different end-to-end neural network architectures for question-answering on tables have been proposed~\cite{neelakantan2015neural,yin2016neural}. Both models show promising results on synthetic datasets, but neither has been evaluated on real data. We implement our own end-to-end neural model (\neural) by generally following both models but deviating when necessary to account for our dataset characteristics.

As a brief description, we encode the question, each column header, and each cell in the table with a character-level \abr{lstm}. We identify three high-level operations based on our dataset characteristics (\emph{select\_column}, \emph{select\_row}, and \emph{select\_cell}) and design modules that perform each of these functions. A module-level soft attention mechanism, effectively a weighted sum of the module scores, decides which module to use given a question.\footnote{We did not design more specific modules to handle arithmetic or aggregation like those of~\newcite{neelakantan2015neural}, although this is a potentially interesting direction for larger datasets.} We also place an additional \abr{lstm} over the question sequence in order to pass information about previous answers and questions to the current time step. Finally, the output of the attention mechanism and the question sequence \abr{lstm} is combined and fed to a binary classifier that, given each cell of the table, decides if the cell is part of the answer to the current question or not.

\begin{figure}[t]
\includegraphics[width=1.0\linewidth]{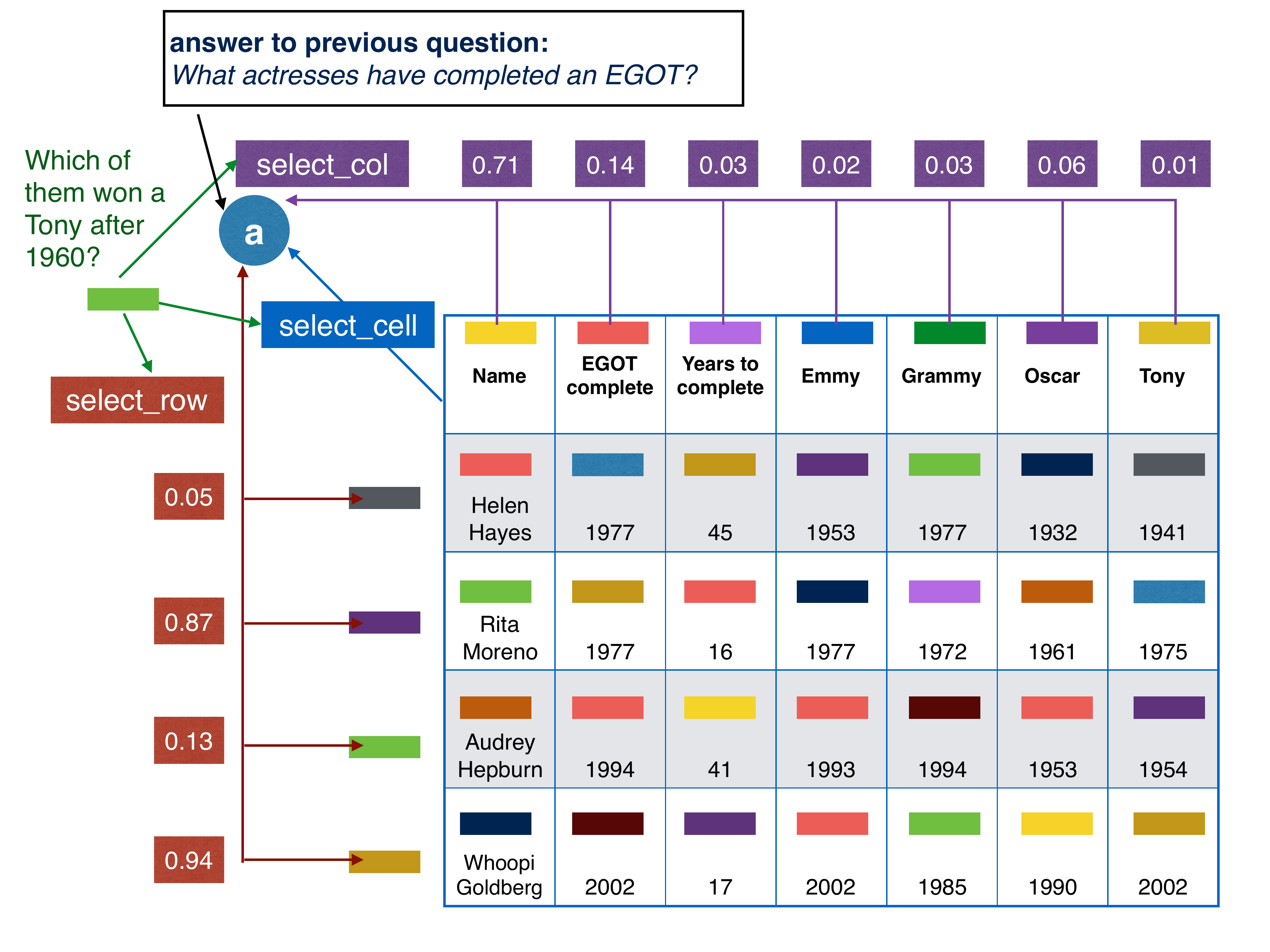}
  \caption{Diagram of \neural\ architecture. Small colored rectangles represent the output of the character-level LSTM decoder. The question, column header, and cell representations are passed to three attentional modules. The output of these modules is combined with the answer predictions for the previous question to yield a final answer prediction for each cell.}
\label{fig:neural}
\end{figure}

Fig.~\ref{fig:neural} shows an example of how the modules in \neural\ work together to answer a given question. In particular, since the question ``\emph{which of them won a Tony after 1960?}" is asking for the names of the actresses, the column selection module places most of its weight on the ``Name'' column, while the row selection module highly weights rows that satisfy the condition ``\emph{Tony after 1960}''. The modules, which take the question and table as input, are merged with an attention mechanism $a$ that also considers the answer to the previous question. A full specification of \neural\ can be found in Appendix~\ref{sec:appendix}.

In contrast to both the neural programmer of~\newcite{neelakantan2015neural} and the neural enquirer of~\newcite{yin2016neural}, we make the simplifying assumption that each question in a sequence can be solved with just a single operation.
Another major difference is that we use a character-level \abr{lstm}, as the training and test vocabulary are radically different.\footnote{Due to the fact that much of our vocabulary (e.g., numbers, entities) is not included in a regular corpus, we suspect that the alternative of leveraging publicly-available word embeddings will not be effective.}

\subsection{Results}

Table~\ref{table:experiments} shows the results of both \fp\ and \neural\ on the test set of \name. We present both the overall accuracy and the accuracy of answers to questions at each position. Although the accuracy of \fp\ on position-1 questions (48.7\%) is much higher than its performance on \wtq\ (37.0\%), the overall accuracy (32.8\%) is still lower, which indicates
that our \name\ dataset remains difficult. In addition, the \neural\ model significantly underperforms \fp, suggesting that it requires more data or more sophisticated architectural design to generalize to all of \name's complexities.

\begin{table}
\centering
\begin{tabular}{cccccc}
\toprule
Model & All & Pos 1 & Pos 2 & Pos 3 & Pos 4 \\
\midrule
\fp & 32.8 & 48.7 & 25.8 & 26.2 & 17.5 \\
\neural & 17.4 & 27.6 & 13.4 & 11.8 & 12.2 \\
\bottomrule
\end{tabular}
\caption{Accuracy of existing systems on our datasets on all questions and questions at all positions within the sequence. }

\label{table:experiments}
\end{table}

\section{Directions for Improving Sequential Question Answering}
\label{sec:adaptations}

In this section, we explore possible directions for improving the system performance in the sequential question answering setting.
We start from investigating different strategies for handling the coreference issues of questions, and then revisited the semantic
matching issue by conducting some error analysis.

\subsection{Adapting Existing Semantic Parsers}

As we observed in Sec.~\ref{sec:experiments}, existing semantic parsers perform suboptimally on \name.
One possible explanation for the suboptimal performance of existing semantic parsers, shown in Table~\ref{table:experiments},
is that questions that contain references to previous questions or answers are not handled properly.
By leveraging \fp\, we propose two ways to deal with this issue: \emph{question rewriting} and \emph{table rewriting}.

\paragraph{Question rewriting:} Take for example the partial sequence ``\emph{what are all the countries that participated in the olympics? which ones won more than two gold medals?}'' Any system that treats these two questions independently of each other has a high likelihood of failing on the second question because ``\emph{ones}'' is not resolved to ``\emph{countries}''. The obvious solution is to apply coreference resolution. However, existing coreference resolution systems struggle at identifying coreferences across two questions, potentially
due to the fact that their training data came from newswire text with few questions.

An alternative approach is to create a set of common referential expressions (e.g., ``ones'', ``them'', ``those'') and replace them with noun phrases from the previous question.
As we do not have ground-truth coreference annotations, we compute upper-bound improvements on question rewriting instead.
That is, we rewrite a reference in a question with all possible noun phrases in the previous question and count the question as correct if any of the rewritten questions are answered correctly. Interestingly, we observe an upper bound improvement of only $\approx$2\% accuracy.

Why is the upper bound so low? An error analysis finds that in many cases, the logical form predicted by \fp\ is wrong even when the referential expression is correctly resolved.
We will discuss this phenomenon more in Sec.~\ref{sec:discussion}, but here we concentrate on another common scenario: the question contains a coreference to the answer of the previous question. If we modify our example sequence to ``\emph{what are all the countries that participated in the olympics in 2012? which ones won more than two gold medals?}'', then simply replacing ``\emph{ones}'' with ``\emph{countries}'' does not resolve the reference.

\paragraph{Table rewriting:} Instead of building a model that can learn to rewrite the second question to ``\emph{which countries won more than two gold medals in 2012}'', or training a semantic parser that can incrementally update the logical form from the previous question as in~\newcite{zettlemoyer2009learning}, we propose to simply rewrite the table based on the first question's answer. %since almost half of the questions in our dataset belong to the \emph{select subset} and \emph{select row} types.
% \todo{SY: I don't understand this sentence...}\textbf{Specifically, we remove all cells that do not belong to the same row(s) as the previous question's answers.}
Specifically, if we know that a particular question is a row or subset selection type, then we also know that its answer must be located in the rows that contain the previous answer. For example, take the second question of the decomposed sequence in Fig.~\ref{fig:decomposition}, which contains a coreference to the answer of the first question (``which \textbf{of them} won a Tony after 1960'') that refers to four actresses. The smallest possible table from which we can still answer this question is one that has four rows (for each of the four actresses) and two columns (``Name'' and ``Tony''). However, identifying the columns necessary to answer each question is often difficult, so we leave this task to the semantic parser and remove only rows (not columns) that do not contain the previous question's answers (see the rewritten table for this example in Fig.~\ref{fig:neural}). In this way, we implicitly resolve the coreference ``of them'', as any rows that do not correspond to actresses are excluded.

% The justification for this method is as follows: if our question requires subset selection, it likely involves solving a condition on an arbitrary column (e.g., ``\emph{which of those countries won five medals?}'' where ``Countries'' and ``Medals'' are columns in the table); if the question is row selection, the current question's answer will come from an arbitrary column. Thus, we cannot remove columns without risking loss of crucial information, but removing irrelevant rows is enough to resolve the inter-question coreferences present in row and subset selection questions.
% \todo{SY: I'm not sure where to check...}\textbf{As an example, see the truncated table in Fig.~\ref{fig:neural}, which is rewritten based on the first question's answer.}

Before rewriting the table, we have to first decide whether the question contains a coreference to the answer or not. We know that we should only rewrite the table for subset and row selection questions. Since we can identify the question type in our dataset based on the coordinates of the answers, we assume that we know which questions should and should not be rewritten and use this information to compute upper bounds for semantic parser improvement with table rewriting. We evaluate five different rewriting policies which vary in their knowledge of both the question type and the correctness of the previous predicted answer:

\begin{enumerate}
	\item \emph{never rewrite} the table
	\item \emph{always rewrite} the table based on the previous predicted answer, regardless of whether table rewriting is applicable to the question
	\item \emph{rewrite row/subset}: rewrite the table based on the previous predicted answer only when table rewriting is applicable (i.e., the question is subset or row selection)
	\item \emph{reference: } same as \emph{rewrite row/subset}, except we only rewrite when we know the previous predicted answer is correct
	\item \emph{upper bound: } same as \emph{rewrite row/subset}, except we rewrite using the previous ground-truth answer instead of the previous predicted answer
\end{enumerate}

% \todo{SY: Need to explain better what the columns mean in the table...}
Table~\ref{table:rewriting} shows the results of running these different rewriting policies on our dev set. The oracle score represents the percentage of questions for which at least one candidate logical form generated by the parser\footnote{The number of candidate parses considered by \fp\ varies and could sometimes be hundreds.} evaluates to the correct answer. The most important takeaway is that accuracy improvements are very small when we rewrite based on the previous predictions. Intuitively, this makes sense: if the parser only gets 30\% accuracy, then 70\% of the time it will be incorrect on the previous question, and rewriting the table based on a wrong answer could make it impossible for the parser to get the right answer (see the lower oracle scores for \emph{always rewrite} and \emph{rewrite row/subset}). Based on these results, table rewriting will only be useful if the base parser's accuracy is high.

\begin{table}
\centering
\begin{tabular}{ccc}
\toprule
Policy & Dev Acc & Dev Oracle \\
\midrule
Never rewrite & 27.7 & 66.6 \\
Always rewrite & 26.9 & 55.3 \\
Rewrite row/subset & 28.2 & 59.8 \\
Reference & 29.2 & 67.3 \\
Upper bound & 37.0 & 71.9 \\
\bottomrule
\end{tabular}
\caption{Dev accuracy of different table rewriting policies; the upper bound represents an almost 10\% absolute improvement that the other policies do not come close to reaching due to the poor baseline performance of \fp. }

\label{table:rewriting}
\end{table}

\subsection{The semantic matching problem}
\label{sec:discussion}

The underwhelming improvements from question and table rewriting force us to re-evaluate our original hypothesis that reference resolution is the main source of complexity in our dataset. 
We take 70 questions from our dev set and manually annotate them with reasons why \fp\ answered them incorrectly. Somewhat surprisingly, we find that only 15 of these errors are due solely to coreferences! The majority of errors are due to wrong logical forms that cannot be corrected by simply resolving a coreference (e.g., the wrong operations are used, or the order of the operations is incorrect).

\begin{figure}[ht]
\includegraphics[width=1.1\linewidth]{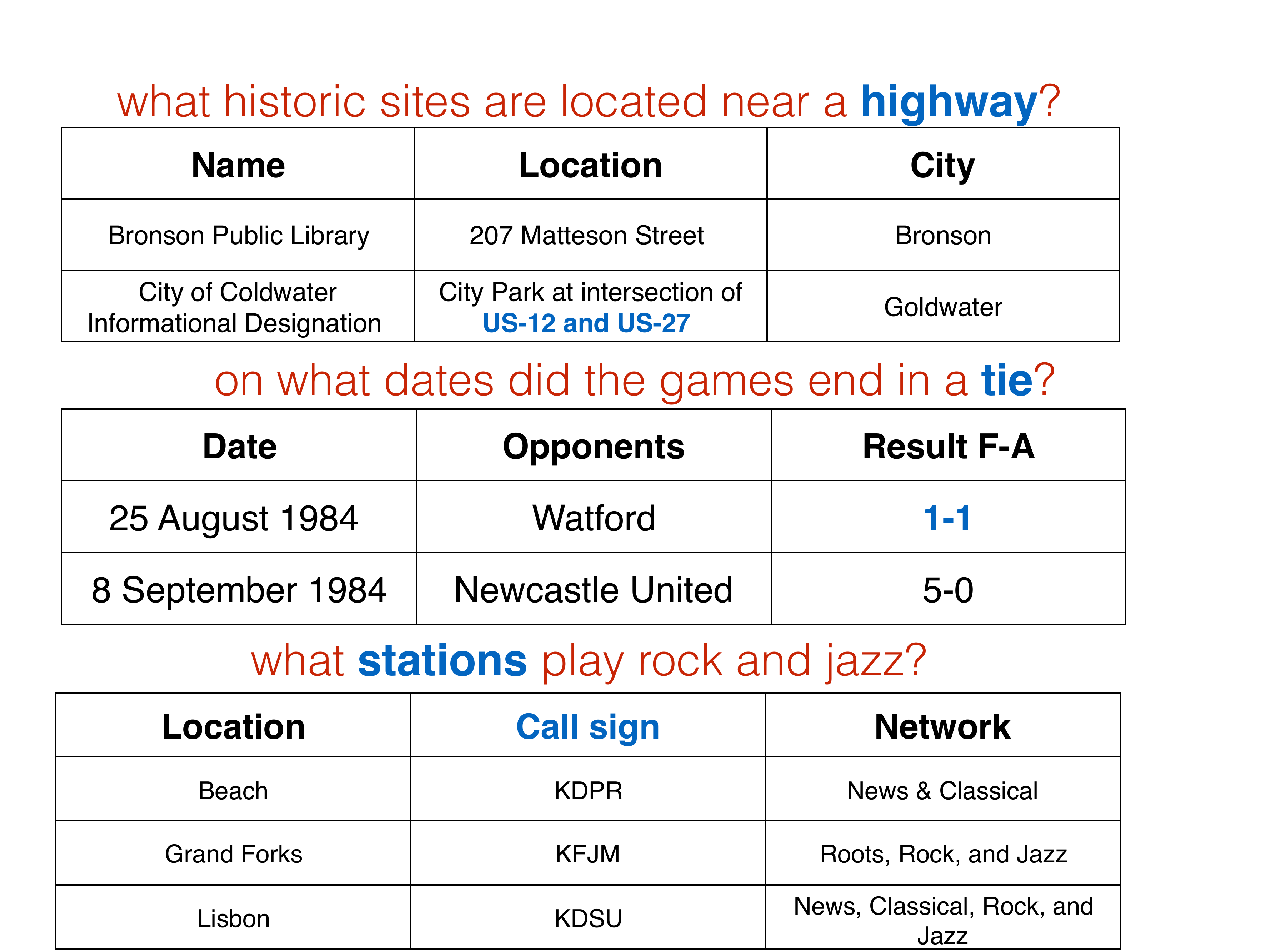}
  \caption{Example mismatches between question and table from \name. Resolving the mismatches requires world knowledge such as highway naming conventions and sports terminology that must be provided externally or learned from a larger corpus.}
\label{fig:matchingerrors}
\end{figure}

When checking these questions in detail, we find that the majority of the errors are due to the \emph{semantic matching problem} --
mismatches between question text and table text.
The error analysis in~\cite{pasupat2015compositional} on the more complicated \wtq\ dataset shows that 25\% of errors are due to these mismatches and an additional 29\% to normalization issues (e.g., an answer cell may contain ``Beijing, China'' but the crowdsourced answer is just ``Beijing''). Because all answers in \name\ are the exact text of cells in the table, we avoid these normalization issues; however, the results in Table~\ref{table:experiments} show that the sequential nature of \name\ makes it equally as difficult as \wtq\ for machines. The examples in Figure~\ref{fig:matchingerrors} suggest that without solving the semantic matching problem, we will not be able to properly take advantage of our question or table rewriting adaptations.

\section{Conclusion}
\label{sec:conclusion}

While most current QA systems assume a single-turn setting, in this work we move towards a more conversational, multi-turn scenario in which systems must rely on prior context to answer the user's current question. To this end, we introduce \name, a dataset that consists of 6,066 unique sequences of inter-related questions about Wikipedia tables, with 17,553 questions-answer pairs in total. To the best of our knowledge, \name\ is the first semantic parsing dataset that addresses sequential question answering, which is a more natural interface for information access.

%\name\ provides great opportunities for research as well. 
The unique setting and task scenario defined in \name\ immediately triggers several interesting research questions, such as
whether the simpler questions make the semantic parsing problem easier and how should a system address the coreferences among questions and answers.
Our preliminary experimental study found that existing systems do not perform well on \name. Moreover, the potential of various kinds of question and table
rewriting strategies for handling coreferences is hindered by semantic matching errors between question text and cells or column headers in the table.
In the near future, we plan to resolve such errors by incorporating large external knowledge sources into semantic parsers. 
Longer-term, we hope that research on \name\ will push towards more interactive settings where systems can ask users 
for clarifications and incorporate user feedback into future predictions.

\bibliographystyle{style/acl2015}
\footnotesize
\bibliography{bib/journal-full,bib/miyyer}

\appendix

\section{Implementation Detail in \neural}
\label{sec:appendix}

We implement each of the three modules in \neural\ with soft attention mechanisms over columns, rows, and cells ($\bvec{m}_{\text{col}}$, $\bvec{m}_{\text{row}}$, and $\bvec{m}_{\text{cell}}$, respectively). While they are functionally similar, each module differs from the others in both inputs and outputs. Before we present the equations defining each module, we introduce some notation: say we have a $r \times c$-dimensional table and an \abr{lstm} that encodes the question into a $d$-dimensional vector $\bvec{q}$. We use the same \abr{lstm} to encode the column headers into a $c \times d$ matrix $\bvec{h}$ and the table cell entries into an $r \times c \times d$ tensor $\bvec{t}_1$.  Similar to the neural enquirer, we add type information to the cell representations by computing a bilinear product with the column headers, $\bvec{t}_{i,j} = \text{ReLu}(\bvec{h}_j \bmat{W}_1 \bvec{t}_{1_{i,j}})$. Before we can implement our modules, we also have to integrate the previous answer predictions ($\bvec{p}_1$ of dimensionality $r \times c$). We use a feed-forward layer to determine how relevant the previous answers are to the current question: $\bvec{p} = \text{ReLu}(\bmat{W}_3 \bvec{p}_1 + \bmat{W}_4 \bvec{q})$. Then, the table representation is updated with the ground-truth previous answers in a simple additive fashion: $\bvec{t} = \bvec{t} + \bvec{p} * \bvec{t}$.

Our modules are defined as follows:

\begin{align}
\label{eq:modules}
\begin{split}
\bvec{m}_{\text{col}} &= \text{softmax}(\bvec{h} \bmat{W}_5 \bvec{q}), \\
\bvec{m}_{\text{row}} &= \sigma((\sum_j \bvec{t}_{i,j})\bmat{W}_6 \bvec{q} ), \\
\bvec{m}_{\text{cell}} &= \sigma(\bvec{t} \bmat{W}_7  \bvec{q})
\end{split}
\end{align}

Note that $\bvec{m}_{\text{col}}$ uses a softmax instead of a sigmoid; most of the questions in \name\ have answers that come from just a single column of the table, so the softmax function's predisposition to select a single input is desirable here. Finally, we compute the final answer predictions $\bvec{a}$ by merging the module outputs with a soft attention mechanism that looks at the question to generate a three-dimensional vector, $\bvec{m}_{\text{att}}$, where each dimension corresponds to the weight for one module.

\begin{align}
\label{eq:attention}
\begin{split}
\bvec{m}_{\text{att}} &= \text{softmax}(\bmat{W}_8\bvec{q}), \\
\bvec{a}_{i,j} &= \sum \bvec{m}_{\text{att}} * [\bvec{m}_{\text{col}_j}; \bvec{m}_{\text{row}_i}; \bvec{m}_{\text{cell}_{i,j}}]
\end{split}
\end{align}

The model parameters are optimized using Adam~\cite{kingma2014adam}; we train for 100 epochs and select the best-performing model on the dev set. We set the dimensionality of our \abr{lstm} hidden state to $d=256$ and the character embedding dimensionality to 100.

\end{document}